# CONFIDENCE CALIBRATION AND RATIONALIZATION FOR LLMS VIA MULTI-AGENT DELIBERATION

**Ruixin Yang**
University of British Columbia

**Dheeraj Rajagopal**
Carnegie Mellon University

**Shirley Anugrah Hayati**
University of Minnesota

**Bin Hu**
University of Minnesota

**Dongyeop Kang**
University of Minnesota

## ABSTRACT

Uncertainty estimation is a significant issue for current large language models (LLMs) that are generally poorly calibrated and over-confident, especially with reinforcement learning from human feedback (RLHF). Unlike humans, whose decisions and confidences not only stem from intrinsic beliefs but can also be adjusted through daily observations, existing calibration methods for LLMs focus on estimating or eliciting individual confidence without taking full advantage of the "Collective Wisdom": the interaction among multiple LLMs that can collectively improve both accuracy and calibration. In this work, we propose *Collaborative Calibration*, a post-hoc training-free calibration strategy that leverages the collaborative and expressive capabilities of multiple tool-augmented LLM agents in a simulated group deliberation process. We demonstrate the effectiveness of *Collaborative Calibration* on generative QA tasks across various domains, showing its potential in harnessing the rationalization of collectively calibrated confidence assessments and improving the reliability of model predictions [1].

## 1 INTRODUCTION

While contemporary large language models (LLMs) have achieved remarkable performance in a variety of tasks ranging from question answering to complex reasoning (Brown et al., 2020; Bubeck et al., 2023), it remains a significant bottleneck for them to produce well-calibrated confidence estimates for their predictions, meaning that their individual confidence is not a reliable indicator of accuracy. Models still often generate hallucinations (Bubeck et al., 2023) or wildly wrong predictions, *unknowingly* and *over-confidently*, which is found to be more evident for models fine-tuned with RLHF (Kadavath et al., 2022; Tian et al., 2023). On the other hand, models can exhibit inconsistencies and lack of confidence, by blindly altering decisions and prioritizing incorrect user opinions (Wei et al., 2023). Such miscalibration is claimed to be even more significant for larger and more capable language models (Kong et al., 2020; Xiong et al., 2023), suggesting the ineffectiveness of model scaling (Kaplan et al., 2020) for mitigating this problem, which poses a great challenge in fostering trust in Human-AI collaboration and in developing reliable real-life applications, especially in high-risk domains.

Although confidence estimation and calibration have been extensively studied in the broader machine learning literature (Gal & Ghahramani, 2016; Guo et al., 2017), previous work in the context of NLP mostly required extensive fine-tuning (Kong et al., 2020) or temperature-based scaling (Guo et al., 2017; Jiang et al., 2021), which can be expensive for LLMs. Recent studies on confidence estimation for black-box LLMs adopted either consistency-based approaches with repeated sampling (Wang et al., 2023b) or verbalization-based elicitation through direct prompting (Lin et al., 2022; Tian et al., 2023), or combined (Xiong et al., 2023). However, results are mixed on whether the elicited confidences are better-calibrated than sample-based estimates or the model's original token probabilities, and no attempt has been made to further improve the interpretability and reliability of such measurement by incorporating the rationalization and collective refinement of individual confidence, leveraging the expressive, self-critical, tool-use and collaborative capabilities of generative LLM agents (Park et al., 2023; Madaan et al., 2024; Schick et al., 2024).

---

[1]Our implementation is publicly available at https://github.com/minnesotanlp/collaborative-calibration.

Inspired by the simple observation that humans can effectively adjust and balance their confidence assessments by weighing agreeing or dissenting opinions from others through group interaction (Silver et al., 2021), we introduce *Collaborative Calibration*, a training-free method for confidence estimation, calibration, and rationalization for LLMs, by simulating a two-stage group deliberation process with multiple LLM agents (outlined in Figure 1). We demonstrate the effectiveness of *Collaborative Calibration* on free-form QA tasks, showing that it can achieve comparable or superior performance on multiple calibration metrics compared to previous methods, without hurting accuracy or generation quality, or at the expense of extensive fine-tuning or parameter-fitting.

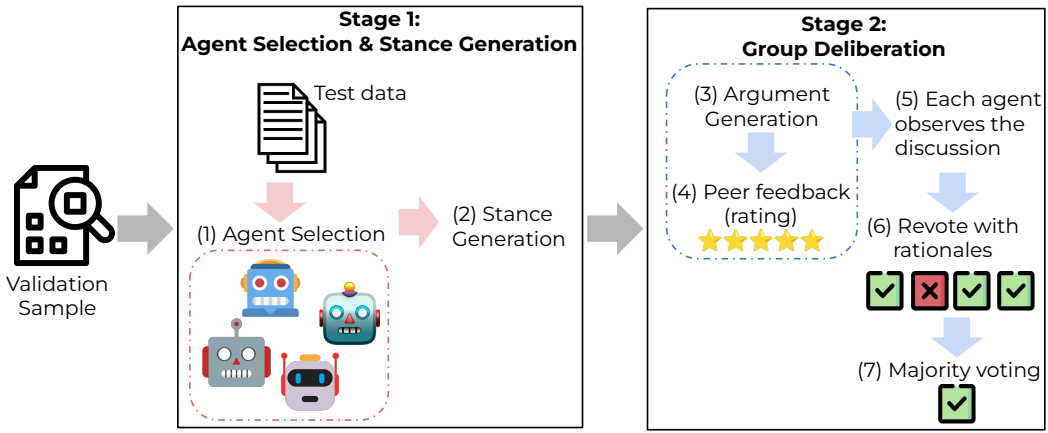

Figure 1: High-level overview of the *Collaborative Calibration* pipeline.

## 2 RELATED WORK

**Confidence Estimation and Calibration for LLMs:** With the rapid advancement of LLMs, there has been an increasing focus on estimating and calibrating their prediction confidence, primarily for classification settings (Jiang et al., 2021; Si et al., 2022; 2023a; Portillo Wightman et al., 2023). For the more challenging yet pertinent tasks of free-form generation with varying answer lengths, Kuhn et al. (2023) introduce an unsupervised entropy-based metric that captures the uncertainty over meanings rather than sequences, and Liu et al. (2024) present a lightweight training method that learns a bias term added to the output logits for better-calibrated confidence estimates. However, these methods require access to internal model structures or the output logits, which are not available for proprietary LLMs. Recent research endeavors have focused on either estimating confidence based on the probability distribution of the most consistent answer from multiple samples (Wang et al., 2023b; Chen et al., 2023c; Lin et al., 2023) or directly eliciting verbalized confidence (Lin et al., 2022; Tian et al., 2023), and found that prompting or fine-tuning with verbalized expression of uncertainty can lead to increased accuracy or better calibration (Mielke et al., 2022; Zhou et al., 2023). Xiong et al. (2023) present a comprehensive analysis of logit-based, verbalized, and consistency-based methods and suggest that a hybrid approach incorporating verbalized confidence into a consistency-based ensemble is more effective for confidence calibration.

**LLM Agent Ensemble:** Augmented with external memory system and tool-use abilities, LLM-powered language agents (Park et al., 2023) display great potential to serve as self-consistent human proxies that can plan by self-reflection and refinement (Shinn et al., 2024; Madaan et al., 2024; Sun et al., 2024). Harnessing the diversity and collaboration strengths of multiple LLM agents, the ensemble and interaction can effectively enhance their capabilities in complex reasoning (Xiong et al., 2023; Chen et al., 2023b; Wang et al., 2023a), instruction following (Jiang et al., 2023b), and value alignment (Liu et al., 2023a). Du et al. (2023) introduce a multi-agent debate framework where multiple LLM instances propose and debate their reasoning processes to reach a consensus, which improves the factuality and performance on arithmetic and strategic reasoning tasks. Similarly, RECONCILE (Chen et al., 2023a) uses different backbone LLMs for debating and a confidence-weighted voting mechanism to obtain the final answer. Our approach differs in that we incorporate self-consistency estimates, and naturally calibrate the verbalized confidence through group deliberation with peer feedback and intermediate rationales for confidence adjustment, rather than manual rescaling in a post-hoc manner. We also incorporate both open and API-based models for the flexibility of supporting both logit-based and black-box confidence estimation. Liu et al. (2023b) propose an inference-time agent selection framework based on aggregated peer ratings over multiple rounds of interactions for a given query. In contrast, our agent selection process requires only the first-round individual confidence estimates at the task level.

## 3 COLLABORATIVE CALIBRATION: CALIBRATING CONFIDENCE VIA MULTI-AGENT DELIBERATION

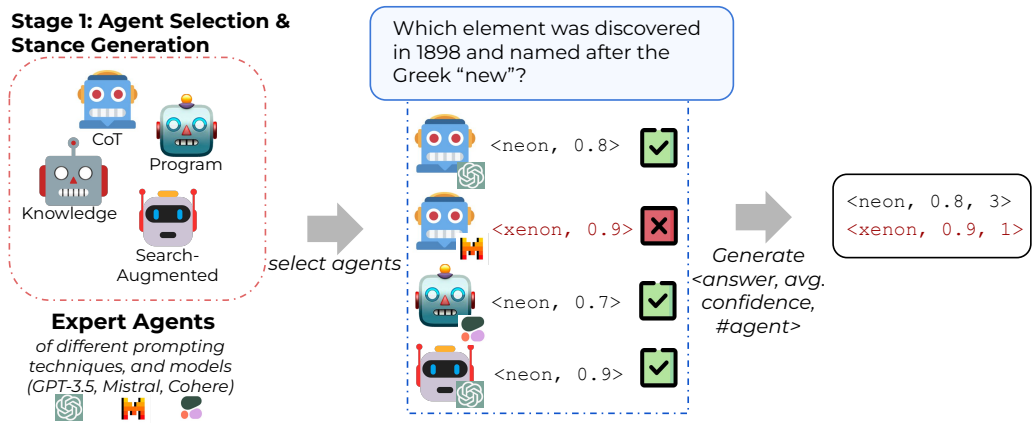

(a) Stage 1 selects the composition of expert agents with suitable prompting techniques or tool-use expertise based on calibration performance on the validation set. For a test query, the selected agents then generate their initial answers, which are clustered into semantically unique stances, with an average confidence and agent count per stance.

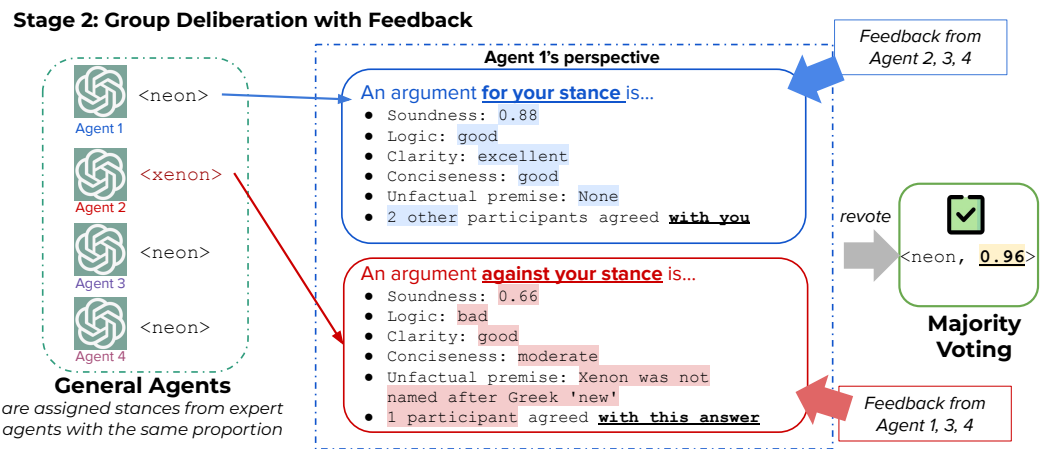

(b) In Stage 2, each general agent provides arguments for its assigned stance, gives feedback on others' arguments, revises its answer, and adjusts the confidence with some rationales. We take a majority vote to get the final, aggregated confidence estimate.

Figure 2: Detailed illustration of the two-stage framework with a specific test example from the SciQ dataset.

## 3.1 AGENT ENSEMBLE AND STANCE GENERATION

To strive for a balance between task accuracy and calibration performance, it is necessary to maintain a certain level of diversity among the agents' initial answers and reasoning paths while allocating the slots wisely so that the most suitable agents for the task can ideally become the majority, which serves as the basis for the ensemble. Inspired by Si et al. (2023b), we initialize four types of "expert agents". Each agent has a different prompting strategy or tool-use expertise: *Chain-of-Thought* (Wei et al., 2022) for multi-hop reasoning tasks, *Program-of-Thoughts* (Chen et al., 2022) for code and arithmetic reasoning tasks, *Search-Augmented Self-Ask* (Press et al., 2022) for factoid reasoning tasks, and GENREAD *prompting* (Yu et al., 2023) for knowledge-intensive reasoning tasks. This initialization is flexible in that any new skill or prompting strategy can be easily added with modularity. As multiple skills might be relevant for an input dataset, we determine the importance ranking of each skill and accordingly allocate expert agents, based on a simple *uncertainty-aware calibration score*, detailed in Appendix A.1. We use Mistral-7B (Jiang et al., 2023a), GPT-3.5-turbo (OpenAI, 2022), and Cohere-Commend (Cohere, 2023) as backbones for the expert agents. For an incoming test question (e.g., *"Which element was discovered in 1898 and named after Greek 'new'?"*), each expert agent performs self-deliberation by executing the corresponding prompting strategy and voting independently for an answer with a confidence estimate. The numerical confidence estimate can be based on the output sequence perplexity (PP) for open models [2], or verbally elicited for black-box models, normalized to $[0, 1]$. Note that this estimate is often unreliable and needs further calibration. We then obtain a set of unique and diverse answers ("stances") by merging semantically equivalent pairs into clusters, using a GPT-3.5 judge following Tian et al. (2023). This constitutes the output of the first stage: semantically unique stances each with a corresponding frequency and aggregated mean confidence.

## 3.2 GROUP DELIBERATION WITH RATIONALES AND FEEDBACK

In the second stage, we initialize another set of "general agents" (with GPT-3.5 backbones and no specialized prompting) to perform rationalization and group deliberation. The diverse stances from Stage 1 are assigned to the general agents as deliberators, proportional to the original answer frequency in Stage 1, thus maintaining potential group consensus or majority voices. However, simply relying on the consensus or majority may not be ideal, as they can sometimes be misleading. This makes the following design necessary — Each agent argues for its assigned stance, producing rationales defending it. This effectively elicits multiple diverse reasoning paths for the ensemble. Agents give ratings and feedback to each argument in terms of logical consistency, factuality, clarity, and conciseness. In particular, the judgment for factuality follows a similar approach to Chain-of-Verification (Dhuliawala et al., 2023), where we ask each agent to self-generate any premise or assumption in the argument under consideration, which could potentially contain hallucination. Therefore, we adopt another search-augmented agent to verify each premise if necessary and highlight any unfactual statement as part of the textual feedback. Each agent is then provided with two rated arguments, one sampled from the affirmative position and one sampled from one of the opposing sides. Observing different arguments with the associated ratings and feedback on factuality, each agent then re-votes their answers $Y_{\text{post}}$, along with rationales $R_{\text{conf}}$ for potential adjustment of its confidence:

$$Y_{\text{post}}, R_{\text{conf}} = M(Y_{\text{prior}}, C_{\text{prior}}, A_p, F_{A_p}, A_n, F_{A_n}) \quad (1)$$

where $Y_{\text{prior}}, C_{\text{prior}}$ are original answers and confidences, $A_p, A_n$ are supporting and opposing arguments, $M$ denotes the underlying model with its parameters, and $F_A$ denotes the summarized peer rating and feedback for argument $A$. In a separate call, each agent gives a posterior confidence estimate based on their final answer and rationales:

$$C_{\text{post}} = \mathbb{P}(Y_{\text{reference}} = Y_{\text{post}} \mid Y_{\text{post}}, R_{\text{conf}}, M) \quad (2)$$

Taking the final majority vote over all $Y_{\text{post}}$ and aggregating the posterior confidences $C_{\text{post}}$, we hypothesize that the final mean confidence estimate will be a better indication of the prediction accuracy, by weighing different voices, evidence, and feedback through deliberation. The intermediate confidence rationales serve as justifications for the final scores, improving the interpretability of potential confidence adjustment.

---

[2] For open models with access to next-token probabilities, the raw confidence for the output sequence $W = (w_1, ..., w_N)$ can be estimated as $\frac{1}{PP(W)} = P(w_1, ..., w_N)^{1/N}$.

Published at ICLR 2024 Workshop on Reliable and Responsible Foundation Models| | GSM8K | | TriviaQA | | SciQ | | AmbigQA | | DateUnd | | Biz-Ethics | |
|---|---|---|---|---|---|---|---|---|---|---|---|---|
| Method | ECE↓ | Brier↓ | ECE↓ | Brier↓ | ECE↓ | Brier↓ | ECE↓ | Brier↓ | ECE↓ | Brier↓ | ECE↓ | Brier↓ |
| Ask4Conf(1S Top-4) | .196* | - | **.054** | .144 | .065 | .209 | - | - | .261* | - | **.124*** | .163 |
| Ask4Conf(2S-CoT) | - | - | .110 | .168 | .323 | .296 | - | - | - | - | - | - |
| Verbalized+Consistency(M=6) | .657 | .620 | .055 | **.050** | .053 | **.094** | .052 | **.098** | .092 | .162 | .141 | .201 |
| Top-K+Self-Random+Avg-Conf | .093 | - | .089† | - | .221† | - | .134† | - | .146 | - | .158 | - |
| **CollabCalibration(M=6)** | **.086** | **.213** | .070 | .062 | **.035** | .129 | **.026** | .126 | **.055** | **.130** | .132 | .203 |

Table 1: We compare *Collaborative Calibration* with previous training-free calibration methods on a variety of tasks using GPT-3.5. (The symbol * denotes results reported by Xiong et al., 2023; † denotes results we reproduce based on the extension of their implementation). Following Tian et al. (2023), each cell in a column is shaded with a gradient from cyan to orange representing varying levels of calibration performance from high to low. The best result of each column is bolded. Our method achieves superior calibration performance in terms of ECE on four of the six tasks, and similar results for other tasks and in terms of Brier Score.

## 4 EXPERIMENTS AND RESULTS

**Metrics:** We use the following metrics for measuring the calibration performance:

*ECE* (Expected Calibration Error, Guo et al., 2017) calculates the average squared error between the estimated confidence and average accuracy within each bin, weighted by the probability that a random sample falls within the bin.

*Brier score* (Brier, 1950) measures the mean squared difference between the predicted probability and the actual outcome, ranging between 0 and 1 where a smaller value is preferred.

**Experimental Setup:** We experiment on six tasks across various domains. Details on the datasets and evaluation methods can be found in Appendix A.2. We compare with multiple previous methods for confidence calibration, including consistency-based ensemble (Wang et al., 2023b), Ask4Conf (Tian et al., 2023), and the best combination of sampling and aggregation approaches reported by Xiong et al. (2023). For fair comparison, we set the ensemble size to 6 for the self-consistency baseline as with our agent group size, and report baseline results evaluated with GPT-3.5. More details on experimental setups are described in Appendix A.2.2.

**Results:** Our main results are reported in Table 1. Compared with the baselines, *Collaborative Calibration* achieves smaller ECE across four of the six tasks, especially for arithmetic and symbolic reasoning tasks (GSM8K, DateUnd) as well as for ambiguity resolution (AmbigQA). On factoid and knowledge-intensive tasks, *Collaborative Calibration* is also able to achieve similar or better results in terms of both ECE and Brier scores. These can be expected as the agents are augmented with programming and search abilities, and can deliberate under uncertainty or ambiguity for reasonable confidence adjustments. Example rationales for the final verbalized confidence can be found in Table 3. Note that for cost-efficiency purposes, the current workflow only adopts search augmentation to a limited degree. Integrating more tool-use modules could further improve calibration performance on knowledge-intensive tasks. Compared with the best set of strategies reported by Xiong et al. (2023), *Collaborative Calibration* can elicit more calibrated confidence scores on the GSM8K and DateUnd datasets, suggesting the effectiveness of additional group interaction and rationalization, beyond simply improving task accuracy. The same pattern is also observed for the comparison with simple consistency-based ensembles, which is shown in detail in Appendix A.3.

## 5 CONCLUSION

In this work, we explore a collaborative approach to elicit, calibrate, and rationalize prediction confidence of LLMs, employing language agents that can be flexible and autonomous in selecting reasoning strategies, using tools, and self-refining with rationales and collective feedback. We demonstrate the flexibility and effectiveness of our method on generative QA tasks across different domains, suggesting possibilities for future work on better leveraging rationales and agent collaboration for improving the reliability of LLM predictions.

## REFERENCES

BIG bench authors. Beyond the imitation game: Quantifying and extrapolating the capabilities of language models. *Transactions on Machine Learning Research*, 2023. ISSN 2835-8856. URL https://openreview.net/forum?id=uyTL5Bvosj.

Glenn W. Brier. Verification of forecasts expressed in terms of probability. *Monthly Weather Review*, 78:1–3, 1950. URL https://api.semanticscholar.org/CorpusID:122906757.

Tom Brown, Benjamin Mann, Nick Ryder, Melanie Subbiah, Jared D Kaplan, Prafulla Dhariwal, Arvind Neelakantan, Pranav Shyam, Girish Sastry, Amanda Askell, Sandhini Agarwal, Ariel Herbert-Voss, Gretchen Krueger, Tom Henighan, Rewon Child, Aditya Ramesh, Daniel Ziegler, Jeffrey Wu, Clemens Winter, Chris Hesse, Mark Chen, Eric Sigler, Mateusz Litwin, Scott Gray, Benjamin Chess, Jack Clark, Christopher Berner, Sam McCandlish, Alec Radford, Ilya Sutskever, and Dario Amodei. Language models are few-shot learners. In H. Larochelle, M. Ranzato, R. Hadsell, M.F. Balcan, and H. Lin (eds.), *Advances in Neural Information Processing Systems*, volume 33, pp. 1877–1901. Curran Associates, Inc., 2020. URL https://proceedings.neurips.cc/paper_files/paper/2020/file/1457c0d6bfcb4967418bfb8ac142f64a-Paper.pdf.

Sébastien Bubeck, Varun Chandrasekaran, Ronen Eldan, Johannes Gehrke, Eric Horvitz, Ece Kamar, Peter Lee, Yin Tat Lee, Yuanzhi Li, Scott Lundberg, et al. Sparks of artificial general intelligence: Early experiments with gpt-4. *arXiv preprint arXiv:2303.12712*, 2023.

Justin Chih-Yao Chen, Swarnadeep Saha, and Mohit Bansal. Reconcile: Round-table conference improves reasoning via consensus among diverse llms. *arXiv preprint arXiv:2309.13007*, 2023a.

Weize Chen, Yusheng Su, Jingwei Zuo, Cheng Yang, Chenfei Yuan, Chi-Min Chan, Heyang Yu, Yaxi Lu, Yi-Hsin Hung, Chen Qian, et al. Agentverse: Facilitating multi-agent collaboration and exploring emergent behaviors. In *The Twelfth International Conference on Learning Representations*, 2023b.

Wenhu Chen, Xueguang Ma, Xinyi Wang, and William W Cohen. Program of thoughts prompting: Disentangling computation from reasoning for numerical reasoning tasks. *arXiv preprint arXiv:2211.12588*, 2022.

Xinyun Chen, Renat Aksitov, Uri Alon, Jie Ren, Kefan Xiao, Pengcheng Yin, Sushant Prakash, Charles Sutton, Xuezhi Wang, and Denny Zhou. Universal self-consistency for large language model generation. *arXiv preprint arXiv:2311.17311*, 2023c.

Karl Cobbe, Vineet Kosaraju, Mohammad Bavarian, Mark Chen, Heewoo Jun, Lukasz Kaiser, Matthias Plappert, Jerry Tworek, Jacob Hilton, Reiichiro Nakano, et al. Training verifiers to solve math word problems. *arXiv preprint arXiv:2110.14168*, 2021.

Cohere. Cohere-command models, 2023. URL https://docs.cohere.com/docs/models.

Shehzaad Dhuliawala, Mojtaba Komeili, Jing Xu, Roberta Raileanu, Xian Li, Asli Celikyilmaz, and Jason Weston. Chain-of-verification reduces hallucination in large language models. *arXiv preprint arXiv:2309.11495*, 2023.

Yilun Du, Shuang Li, Antonio Torralba, Joshua B Tenenbaum, and Igor Mordatch. Improving factuality and reasoning in language models through multiagent debate. *arXiv preprint arXiv:2305.14325*, 2023.

Yarin Gal and Zoubin Ghahramani. Dropout as a bayesian approximation: Representing model uncertainty in deep learning. In Maria Florina Balcan and Kilian Q. Weinberger (eds.), *Proceedings of The 33rd International Conference on Machine Learning*, volume 48 of *Proceedings of Machine Learning Research*, pp. 1050–1059, New York, New York, USA, 20–22 Jun 2016. PMLR. URL https://proceedings.mlr.press/v48/gal16.html.

Chuan Guo, Geoff Pleiss, Yu Sun, and Kilian Q. Weinberger. On calibration of modern neural networks. In Doina Precup and Yee Whye Teh (eds.), *Proceedings of the 34th International Conference on Machine Learning*, volume 70 of *Proceedings of Machine Learning Research*, pp. 1321–1330. PMLR, 06–11 Aug 2017. URL https://proceedings.mlr.press/v70/guo17a.html.

Dan Hendrycks, Collin Burns, Steven Basart, Andy Zou, Mantas Mazeika, Dawn Song, and Jacob Steinhardt. Measuring massive multitask language understanding. *Proceedings of the International Conference on Learning Representations (ICLR)*, 2021.

Albert Q Jiang, Alexandre Sablayrolles, Arthur Mensch, Chris Bamford, Devendra Singh Chaplot, Diego de las Casas, Florian Bressand, Gianna Lengyel, Guillaume Lample, Lucile Saulnier, et al. Mistral 7b. *arXiv preprint arXiv:2310.06825*, 2023a.

Dongfu Jiang, Xiang Ren, and Bill Yuchen Lin. Llm-blender: Ensembling large language models with pairwise comparison and generative fusion. In *Proceedings of the 61th Annual Meeting of the Association for Computational Linguistics (ACL 2023)*, 2023b.

Zhengbao Jiang, Jun Araki, Haibo Ding, and Graham Neubig. How can we know when language models know? on the calibration of language models for question answering. *Transactions of the Association for Computational Linguistics*, 9:962–977, 2021. doi: 10.1162/tacl_a_00407. URL https://aclanthology.org/2021.tacl-1.57.

Mandar Joshi, Eunsol Choi, Daniel Weld, and Luke Zettlemoyer. TriviaQA: A large scale distantly supervised challenge dataset for reading comprehension. In Regina Barzilay and Min-Yen Kan (eds.), *Proceedings of the 55th Annual Meeting of the Association for Computational Linguistics (Volume 1: Long Papers)*, pp. 1601–1611, Vancouver, Canada, July 2017. Association for Computational Linguistics. doi: 10.18653/v1/P17-1147. URL https://aclanthology.org/P17-1147.

Saurav Kadavath, Tom Conerly, Amanda Askell, Tom Henighan, Dawn Drain, Ethan Perez, Nicholas Schiefer, Zac Hatfield-Dodds, Nova DasSarma, Eli Tran-Johnson, et al. Language models (mostly) know what they know. *arXiv preprint arXiv:2207.05221*, 2022.

Jared Kaplan, Sam McCandlish, Tom Henighan, Tom B Brown, Benjamin Chess, Rewon Child, Scott Gray, Alec Radford, Jeffrey Wu, and Dario Amodei. Scaling laws for neural language models. *arXiv preprint arXiv:2001.08361*, 2020.

Lingkai Kong, Haoming Jiang, Yuchen Zhuang, Jie Lyu, Tuo Zhao, and Chao Zhang. Calibrated language model fine-tuning for in- and out-of-distribution data. In Bonnie Webber, Trevor Cohn, Yulan He, and Yang Liu (eds.), *Proceedings of the 2020 Conference on Empirical Methods in Natural Language Processing (EMNLP)*, pp. 1326–1340, Online, November 2020. Association for Computational Linguistics. doi: 10.18653/v1/2020.emnlp-main.102. URL https://aclanthology.org/2020.emnlp-main.102.

Lorenz Kuhn, Yarin Gal, and Sebastian Farquhar. Semantic uncertainty: Linguistic invariances for uncertainty estimation in natural language generation. In *The Eleventh International Conference on Learning Representations*, 2023. URL https://openreview.net/forum?id=VD-AYtP0dve.

Stephanie Lin, Jacob Hilton, and Owain Evans. Teaching models to express their uncertainty in words. *Transactions on Machine Learning Research*, 2022. ISSN 2835-8856. URL https://openreview.net/forum?id=8s8K2UZGTZ.

Zhen Lin, Shubhendu Trivedi, and Jimeng Sun. Generating with confidence: Uncertainty quantification for black-box large language models. *arXiv preprint arXiv:2305.19187*, 2023.

Ruibo Liu, Ruixin Yang, Chenyan Jia, Ge Zhang, Diyi Yang, and Soroush Vosoughi. Training socially aligned language models on simulated social interactions. In *The Twelfth International Conference on Learning Representations*, 2023a.

Xin Liu, Muhammad Khalifa, and Lu Wang. Litcab: Lightweight language model calibration over short- and long-form responses. In *The Twelfth International Conference on Learning Representations*, 2024. URL https://openreview.net/forum?id=jH67LHVOIO.

Zijun Liu, Yanzhe Zhang, Peng Li, Yang Liu, and Diyi Yang. Dynamic llm-agent network: An llm-agent collaboration framework with agent team optimization. *arXiv preprint arXiv:2310.02170*, 2023b.

Aman Madaan, Niket Tandon, Prakhar Gupta, Skyler Hallinan, Luyu Gao, Sarah Wiegreffe, Uri Alon, Nouha Dziri, Shrimai Prabhumoye, Yiming Yang, et al. Self-refine: Iterative refinement with self-feedback. *Advances in Neural Information Processing Systems*, 36, 2024.

Sabrina J. Mielke, Arthur Szlam, Emily Dinan, and Y-Lan Boureau. Reducing conversational agents' overconfidence through linguistic calibration. *Transactions of the Association for Computational Linguistics*, 10:857–872, 2022. doi: 10.1162/tacl_a_00494. URL https://aclanthology.org/2022.tacl-1.50.

Sewon Min, Julian Michael, Hannaneh Hajishirzi, and Luke Zettlemoyer. AmbigQA: Answering ambiguous open-domain questions. In Bonnie Webber, Trevor Cohn, Yulan He, and Yang Liu (eds.), *Proceedings of the 2020 Conference on Empirical Methods in Natural Language Processing (EMNLP)*, pp. 5783–5797, Online, November 2020. Association for Computational Linguistics. doi: 10.18653/v1/2020.emnlp-main.466. URL https://aclanthology.org/2020.emnlp-main.466.

OpenAI. Chatgpt blog, 2022. URL https://openai.com/blog/chatgpt.

Joon Sung Park, Joseph C. O'Brien, Carrie J. Cai, Meredith Ringel Morris, Percy Liang, and Michael S. Bernstein. Generative agents: Interactive simulacra of human behavior. In *In the 36th Annual ACM Symposium on User Interface Software and Technology (UIST '23)*, UIST '23, New York, NY, USA, 2023. Association for Computing Machinery.

Gwenyth Portillo Wightman, Alexandra Delucia, and Mark Dredze. Strength in numbers: Estimating confidence of large language models by prompt agreement. In Anaelia Ovalle, Kai-Wei Chang, Ninareh Mehrabi, Yada Pruksachatkun, Aram Galystan, Jwala Dhamala, Apurv Verma, Trista Cao, Anoop Kumar, and Rahul Gupta (eds.), *Proceedings of the 3rd Workshop on Trustworthy Natural Language Processing (TrustNLP 2023)*, pp. 326–362, Toronto, Canada, July 2023. Association for Computational Linguistics. doi: 10.18653/v1/2023.trustnlp-1.28. URL https://aclanthology.org/2023.trustnlp-1.28.

Ofir Press, Muru Zhang, Sewon Min, Ludwig Schmidt, Noah A Smith, and Mike Lewis. Measuring and narrowing the compositionality gap in language models. *arXiv preprint arXiv:2210.03350*, 2022.

Timo Schick, Jane Dwivedi-Yu, Roberto Dessì, Roberta Raileanu, Maria Lomeli, Eric Hambro, Luke Zettlemoyer, Nicola Cancedda, and Thomas Scialom. Toolformer: Language models can teach themselves to use tools. *Advances in Neural Information Processing Systems*, 36, 2024.

Noah Shinn, Federico Cassano, Ashwin Gopinath, Karthik Narasimhan, and Shunyu Yao. Reflexion: Language agents with verbal reinforcement learning. *Advances in Neural Information Processing Systems*, 36, 2024.

Chenglei Si, Chen Zhao, Sewon Min, and Jordan Boyd-Graber. Re-examining calibration: The case of question answering. In Yoav Goldberg, Zornitsa Kozareva, and Yue Zhang (eds.), *Findings of the Association for Computational Linguistics: EMNLP 2022*, pp. 2814–2829, Abu Dhabi, United Arab Emirates, December 2022. Association for Computational Linguistics. doi: 10.18653/v1/2022.findings-emnlp.204. URL https://aclanthology.org/2022.findings-emnlp.204.

Chenglei Si, Zhe Gan, Zhengyuan Yang, Shuohang Wang, Jianfeng Wang, Jordan Lee Boyd-Graber, and Lijuan Wang. Prompting GPT-3 to be reliable. In *The Eleventh International Conference on Learning Representations*, 2023a. URL https://openreview.net/forum?id=98p5x51L5af.

Chenglei Si, Weijia Shi, Chen Zhao, Luke Zettlemoyer, and Jordan Boyd-Graber. Getting more out of mixture of language model reasoning experts. In *Findings of the Association for Computational Linguistics: EMNLP 2023*, pp. 8234–8249, 2023b.

Ike Silver, Barbara A. Mellers, and Philip E. Tetlock. Wise teamwork: Collective confidence calibration predicts the effectiveness of group discussion. *Journal of Experimental Social Psychology*, 96:104157, 2021. ISSN 0022-1031. doi: https://doi.org/10.1016/j.jesp.2021.104157. URL https://www.sciencedirect.com/science/article/pii/S0022103121000603.

Haotian Sun, Yuchen Zhuang, Lingkai Kong, Bo Dai, and Chao Zhang. Adaplanner: Adaptive planning from feedback with language models. *Advances in Neural Information Processing Systems*, 36, 2024.

Katherine Tian, Eric Mitchell, Allan Zhou, Archit Sharma, Rafael Rafailov, Huaxiu Yao, Chelsea Finn, and Christopher D Manning. Just ask for calibration: Strategies for eliciting calibrated confidence scores from language models fine-tuned with human feedback. *arXiv preprint arXiv:2305.14975*, 2023.

Kuan Wang, Yadong Lu, Michael Santacroce, Yeyun Gong, Chao Zhang, and Yelong Shen. Adapting llm agents through communication. *arXiv preprint arXiv:2310.01444*, 2023a.

Xuezhi Wang, Jason Wei, Dale Schuurmans, Quoc V Le, Ed H. Chi, Sharan Narang, Aakanksha Chowdhery, and Denny Zhou. Self-consistency improves chain of thought reasoning in language models. In *The Eleventh International Conference on Learning Representations*, 2023b. URL https://openreview.net/forum?id=1PL1NIMMrw.

Jason Wei, Xuezhi Wang, Dale Schuurmans, Maarten Bosma, brian ichter, Fei Xia, Ed Chi, Quoc V Le, and Denny Zhou. Chain-of-thought prompting elicits reasoning in large language models. In S. Koyejo, S. Mohamed, A. Agarwal, D. Belgrave, K. Cho, and A. Oh (eds.), *Advances in Neural Information Processing Systems*, volume 35, pp. 24824–24837. Curran Associates, Inc., 2022. URL https://proceedings.neurips.cc/paper_files/paper/2022/file/9d5609613524ecf4f15af0f7b31abca4-Paper-Conference.pdf.

Jerry Wei, Da Huang, Yifeng Lu, Denny Zhou, and Quoc V Le. Simple synthetic data reduces sycophancy in large language models. *arXiv preprint arXiv:2308.03958*, 2023.

Johannes Welbl, Nelson F. Liu, and Matt Gardner. Crowdsourcing multiple choice science questions. In Leon Derczynski, Wei Xu, Alan Ritter, and Tim Baldwin (eds.), *Proceedings of the 3rd Workshop on Noisy User-generated Text*, pp. 94–106, Copenhagen, Denmark, September 2017. Association for Computational Linguistics. doi: 10.18653/v1/W17-4413. URL https://aclanthology.org/W17-4413.

Kai Xiong, Xiao Ding, Yixin Cao, Ting Liu, and Bing Qin. Examining the inter-consistency of large language models: An in-depth analysis via debate. Association for Computational Linguistics, 2023.

Wenhao Yu, Dan Iter, Shuohang Wang, Yichong Xu, Mingxuan Ju, Soumya Sanyal, Chenguang Zhu, Michael Zeng, and Meng Jiang. Generate rather than retrieve: Large language models are strong context generators. In *The Eleventh International Conference on Learning Representations*, 2023. URL https://openreview.net/forum?id=fB0hRu9GZUS.

Kaitlyn Zhou, Dan Jurafsky, and Tatsunori Hashimoto. Navigating the grey area: How expressions of uncertainty and overconfidence affect language models. In Houda Bouamor, Juan Pino, and Kalika Bali (eds.), *Proceedings of the 2023 Conference on Empirical Methods in Natural Language Processing*, pp. 5506–5524, Singapore, December 2023. Association for Computational Linguistics. doi: 10.18653/v1/2023.emnlp-main.335. URL https://aclanthology.org/2023.emnlp-main.335.

## A APPENDIX

### A.1 DETAILS ON AGENT SELECTION

To determine the composition of expert agents used in Stage 1, we rank the relevance of each of the four pre-specified skills based on calibration performance on the validation set. We first sample $m$ examples from the development split and initialize one expert agent per skill and model. For an example query $j$, each candidate agent independently gives initial answers $a_{i,j}$ and uncalibrated confidence estimates $c_{i,j}$ (logit-based for open models and verbal-based for black-box models, normalized to $[0,1]$), where $i = 1,...,4$ and $j = 1,...,m$. To account for the validation performance and penalize abstention or incorrect predictions with high confidence, we multiply $c_{i,j}$ by one of $\{-1, 0, 1\}$ to get the *uncertainty-aware calibration score* in $[-1, 1]$:

$$c'_{i,j} = \begin{cases} 0 & a_{i,j} \text{ is "Abstain"} \\ (2 \cdot \mathbf{1}_{a_{i,j} \in Y_{\text{ref}}} - 1) \cdot c_{i,j} & \text{otherwise} \end{cases} \quad (3)$$

The intuition behind this score is to manually boost confidence given a correct answer, and diminish it otherwise. Aggregating confidence over all examples for each agent and filtering out those below a certain threshold $\tau$ (set to 0.2 in our experiments), we get a vector of adjusted and filtered confidences:

$$c'_i = \tfrac{1}{m} \sum_{j=1}^{m} c'_{i,j}, \text{ and } \mathbf{c} = vec(\{c'_i : c'_i \geq \tau\}) \quad (4)$$

For a total of $N$ slots, we then allocate roughly $\lfloor N * \text{Softmax}(\mathbf{c}) \rfloor$ slots to different agent types.

## A.2 DETAILS ON EXPERIMENT SETUP

### A.2.1 DATASETS

We experiment on various tasks including GSM8K (Cobbe et al., 2021) for arithmetic reasoning, TriviaQA (Joshi et al., 2017), and SciQ (Welbl et al., 2017) for factoid and knowledge-intensive questions, AmbigQA (Min et al., 2020) for reasoning under ambiguity, and Date Understanding (DateUnd) from BigBench (bench authors, 2023) for symbolic reasoning, as well as the Business Ethics (Biz-Ethics) dataset from MMLU (Hendrycks et al., 2021) for questions requiring ethical knowledge. We randomly sample 300 questions for each task from the test or validation split for our experiment, except for the smaller Biz-Ethics dataset where we use all the examples.

### A.2.2 EVALUATION METHODS

We explore the optimal number of agents and the number of feedback per argument, which are set to 6 and 2 respectively. We experiment with an ensemble of Mistral-7B (Jiang et al., 2023a), GPT-3.5-turbo (OpenAI, 2022), and Cohere-Commend (Cohere, 2023) as backbones for Stage 1 expert agents, with two dynamically selected agent types for each model. For agent ensemble, the dynamic agent selection workflow selects CoT and PoT agents for arithmetic tasks and CoT and Knowledge agents for other tasks. Both Cohere and Mistral-7B are less performant than GPT-3.5 in task accuracy, while arguably better calibrated (Cohere) or providing more reliable measures with length-normalized sequence logits (Mistral). By combining their respective strengths, we hypothesize that it would bring better calibration without hurting task accuracy, compared with the poorly calibrated zero-shot settings with solely GPT-3.5 and a simple self-consistency ensemble.

## A.3 DETAILED RESULTS

We present in Figure 3 a detailed comparison between the post-deliberation confidence from *Collaborative Calibration* and the aggregated confidence from the self-consistency ensemble (with the same group size) on GSM8K, SciQ, and DateUnd. As shown, simple ensemble methods can still produce highly concentrated and mostly overconfident estimates (top row), whereas *Collaborative Calibration* yields a more diverse confidence distribution, with the majority cases (denoted by the alpha scale of the bar colors) aligned well with the diagonal, which suggests better calibration performance.

We also perform ablation on the effectiveness of the group deliberation in Stage 2. Figure 4 shows the calibration performance before deliberation (i.e. output from the Stage 1 ensemble) and after, which displays a significant decrease in ECE and Brier scores.

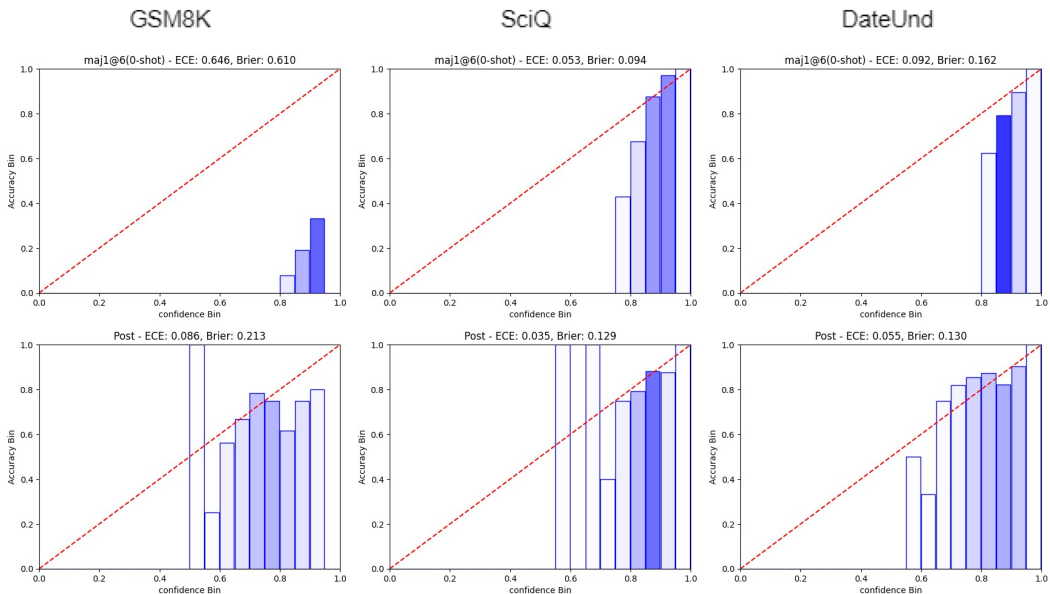

Figure 3: Reliability diagrams comparing vanilla verbalized confidence + Self-consistency (M=6) and our Collaborative Calibration with an ensemble of 6 agents on GSM8K, SciQ, and DateUnd

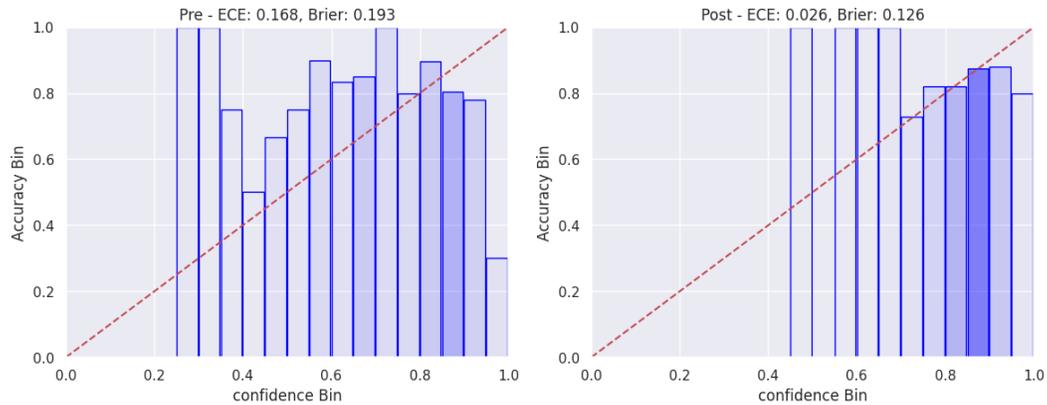

Figure 4: Reliability diagrams comparing calibration performance before and after Stage 2 (group deliberation) on TriviaQA

### A.4  PROMPT TEMPLATES AND EXAMPLE OUTPUT

We follow the prompt templates from the official implementations of CoT (Wei et al., 2022), PoT (Chen et al., 2022), and Knowledge prompting (Yu et al., 2023). For cost considerations, we adopt the zero-shot setting for each strategy (e.g. *"Let's think/write Python programs step-by-step"* without providing a demonstration) and resort to search verification only when the model is unsure about a generated premise. Other prompt templates, including the prompts for argument generation, feedback, and final confidence can be found in Table 2.

Table 2: Prompt templates used across the two stages in *Collaborative Calibration*.

| | Template |
|---|---|
| **Stance generation** | State your answer (as short as possible, in one or a few words), then rate the level of ambiguity in the input query (a float from 0 to 1); rate the level of complexity of the input query (a float from 0 to 1); rate your level of ability for solving the input query (a float from 0 to 1); Note that your uncertainty on the correctness of your answer is affected by input ambiguity, task complexity, and your own knowledge and abilities. Based on this, give a float (between 0 to 1) indicating your overall confidence on how likely that your answer is correct. Follow this format: "Answer:\<answer\> Confidence:\<confidence\>" |
| **Argument generation** | You are participating in a debate on the question: "${QUERY}" Your assigned stance on the question is "${STANCE}" Generate some arguments or evidence (no more than three sentences) on why your assigned stance is correct. If the question is ambiguous, address the assumptions or interpretations associated with your assigned stance. Be concise! Exclude anything irrelevant or unhelpful in terms of supporting the stance! Argument: |
| **Argument rating** | Here is an argument "${ARGUMENT}" for the stance "${STANCE}". Note in the earlier debate, you supported the answer corresponding to this argument. Evaluate how good the argument is regarding logical consistency, clarity, and conciseness. For each of the three aspects, choose one of 'bad', 'modest', 'good', and 'excellent' as your rating. Do NOT provide any reasoning. Follow this format: "Consistency: \<rating\>, Clarity:\<rating\>, Conciseness:\<rating\>" |
| **Confidence rationale generation** | Given the question: "${QUERY}", your original answer is "${STANCE}", with a confidence score of ${ORIGINAL-CONFIDENCE} Here are some new observations: ''' An argument from the opposing side is "${ARGUMENT-AGAINST}", which received the following rating and feedback from other deliberators: "${FEEDBACK-AGAINST}" Note that ${NUMBER-AGAINST} people disagreed with you. An argument supporting your original answer is "${ARGUMENT-FOR}", which received the following rating and feedback from other deliberators: "${FEEDBACK-SUPPORTING}" ''' Note that ${NUMBER-SUPPORTING} other people also agreed with you. Give your final answer to the question (as short as possible). Considering your original belief, group consensus, and new observations, and weighing arguments from multiple sides (including your own), give rationales for whether you would adjust your original confidence score. Follow this format:"Answer: \<answer\> Rationales: \<rationales\>" |
| **Final confidence elicitation** | Recall your original confidence for your answer is ${ORIGINAL-CONFIDENCE}. Given the rationale: "${CONFIDENCE-RATIONALE}", provide your final confidence score (a float from 0 to 1). Follow this format: "Confidence: \<confidence\>" |

Table 3: Examples for rationales generated by the deliberator agents for their final confidence adjustments, on sampled questions from SciQ and GPT-3.5-turbo as the backbone model.

| Sample Rationales |
|---|
| The group consensus and new observations indicate that the behavior of a compound is influenced by multiple factors, including its chemical structure, physical properties, and environmental conditions. Therefore, adjusting the original confidence score to account for the multiple contributing factors would be appropriate. |
| My original belief was based on my understanding of fungal reproduction, and the group consensus reinforces the confidence in my answer. The new observations of unanimous agreement further support the accuracy of the answer. Therefore, I would not adjust my original confidence score. |
| The new observations and feedback from other deliberators support the argument that anions play a crucial role in balancing the charge in ionic compounds. Additionally, the majority of the group disagreed with the original answer and supported the opposing argument. Therefore, the original confidence score should be adjusted to reflect the new information and group consensus. |
| The argument supporting the original answer received a high soundness score and agreement from the majority of deliberators. The argument against the original answer also received a high soundness score, but with less agreement. Considering the balanced feedback and group consensus, the original confidence score of 0.60 can be adjusted upwards to reflect the stronger supporting argument and group alignment. |